\def\year{2018}\relax
\documentclass[letterpaper]{article} 
\usepackage{aaai18}  
\usepackage{times}  
\usepackage{amsmath}
\usepackage{float} 
\usepackage{booktabs} 
\usepackage{todonotes}
\usepackage{helvet}  
\usepackage{courier}  
\usepackage{url}  
\usepackage{graphicx}  
\usepackage{verbatim} 
\usepackage{balance} 
\title{Always Lurking: Understanding and Mitigating Bias in Online Human Trafficking Detection}

\author{
Kyle Hundman \\
kyle.a.hundman@jpl.nasa.gov \\
NASA Jet Propulsion Laboratory \\
California Institute of Technology \\
Pasadena, CA 91109 USA \\
\And 
Thamme Gowda \thanks{This work was done when the author was at the NASA Jet Propulsion Laboratory, Pasadena, CA, USA.} \\
tg@isi.edu\\
      Information Sciences Institute \\
                University of Southern California \\
                Marina Del Rey, CA 90292, USA \\
\AND 
Mayank Kejriwal \\ 
kejriwal@isi.edu \\
Information Sciences Institute \\
		        University of Southern California \\
       		   Marina Del Rey, CA 90292, USA \\
\And
Benedikt Boecking \\
boecking@cmu.edu \\
      Carnegie Mellon University \\
       			Pittsburgh, PA 15213, USA \\
}
\begin{document}

\maketitle

\begin{abstract}

Web-based human trafficking activity has increased in recent years but it remains sparsely dispersed among escort advertisements and difficult to identify due to its often-latent nature. The use of intelligent systems to detect trafficking can thus have a direct impact on investigative resource allocation and decision-making, and, more broadly, help curb a widespread social problem. Trafficking detection involves assigning a normalized score to a set of escort advertisements crawled from the Web -- a higher score indicates a greater risk of trafficking-related (involuntary) activities. In this paper, we define and study the problem of trafficking detection and present a trafficking detection pipeline architecture developed over three years of research within the DARPA Memex program. Drawing on multi-institutional data, systems, and experiences collected during this time, we also conduct post hoc bias analyses and present a bias mitigation plan. Our findings show that, while automatic trafficking detection is an important application of AI for social good, it also provides cautionary lessons for deploying predictive machine learning algorithms without appropriate de-biasing. This ultimately led to integration of an interpretable solution into a search system that contains over 100 million advertisements and is used by over 200 law enforcement agencies to investigate leads.

\end{abstract}

\section{Introduction}
\label{sec:intro}


Human trafficking has seen increasing media attention and government focus in recent years due to its pervasiveness and insidious nature \cite{austin2017human}. It is also characterized by a significant \emph{Web presence}, with traffickers often advertising their victims on public platforms such as \url{backpage.com} \cite{szekely2015:iswc}. Forums and review sites also contain discussions by `clients' about (potentially trafficked) escorts, and other aspects of their experiences, such as the youth of, and services provided by, the escort.


The high prevalence of online sex advertisements (ads) and reviews, even on the Open Web, was a motivating factor in the creation of the DARPA Memex\footnote{\url{https://www.darpa.mil/program/memex}} program, under which this work was funded and conducted over a period of three years. Memex was designed to advance the state-of-the-art in building domain-specific search systems over massive Web corpora, especially in difficult domains like human trafficking. Various Memex-funded systems can be integrated to build \emph{end-to-end} domain-specific search systems, starting from \emph{domain modeling} and \emph{discovery} (including crawling the Web for relevant pages), knowledge graph construction, machine learning and information retrieval \cite{szekely2015:iswc}, \cite{szekely2017}, \cite{ache}, \cite{deepdive}.

An important \emph{inferential} problem that needs to be addressed at scale in this pipeline is to detect potential trafficking activity by assigning a \emph{risk score} to a set of advertisements, usually collected by an investigative expert like a law enforcement official. In the simplest case, the risk score is a binary flag, with 1 indicating that the ads in the set warrant further trafficking-related investigation. Intuitively, a set represents an informal version of a `case study' that, for reasons grounded in real-world activities like tip-offs from contacts in the field, arrest records or exploratory search, has come to the attention of an official. While a single ad is often not useful by itself, intriguingly, when considered in aggregate, even a small set of ads in the case study can provide \emph{subtle} clues indicating trafficking, rather than voluntary escort activities. For example, there may be evidence in one of the ads that an escort is \emph{underage} or is advertising sex services that are \emph{risky} and unusual relative to the domain. There may also be evidence of \emph{movement} between cities, or in the case of brothels often fronting as Asian massage parlors, \emph{ethnicity}-related clues. 

The problem of trafficking detection, even by a human carefully analyzing the case study, is further compounded by the fact that ads in such case studies tend to be related only \emph{latently}, and the relation itself can be subtle. In most cases, the assumption in identifying trafficking-related case studies is that escorts represented in the case study ads are being trafficked in a similar \emph{context}, i.e. by a \emph{single} individual or an organization. Precisely identifying such contexts, and evidence backing the contexts, can be used to inform investigative decision making, alleviate the cognitive burden and significantly preserve the limited resources of both law enforcement and state district attorneys, who are often tasked with prosecuting trafficking-related cases. Given that the Memex program has already scraped many millions of sex advertisements on the Web for investigative purposes \cite{szekely2017}, \emph{automatic detection} methods, ideally with \emph{interpretation}, can serve an important, socially beneficial function. At the same time, because the latest state-of-the-art methods in text classification use complex machine learning models like deep neural networks \cite{deeptext}, with strong (and not very well understood) dependencies on the input data, it is important to understand the \emph{biases} and limitations of such methods. 

Systematically understanding the tradeoff described above, between building a system that can serve as an important example of \emph{AI for Social Good}, and ensuring that the system is \emph{fit} for use in a real world, requires a collaborative, socio-technological exchange. The Memex program provided such an exchange through a series of competitions and \emph{quarterly progress reviews} (QPRs) when participants would regularly gather for myriad purposes such as system evaluation, strategic collaborations, and detailed meetings with potential real-world users of the system such as law enforcement. The QPRs led to a rich trove of insights around trafficking detection, both as a \emph{technical} and an \emph{investigative field} problem. In this paper, we use these insights to present the trafficking detection problem in depth for the broader research community, solutions explored and developed over the course of three years, corresponding lessons learned, eventual (and ongoing) integration into a comprehensive search system, and continuing impact. Specific contributions are listed below.

{\bf Contributions.}
The main contributions of this work are described as follows. We describe the trafficking detection problem and the motivations for devising automatic trafficking detection methods. We present an \emph{architectural overview} of the approach that was first developed (before bias analysis and mitigation), followed by a \emph{bias mitigation plan} that was put in place and continues to be implemented. We describe the lessons learned over three years of research from this exercise, and the changes that were made to the system in response to real-world users. The most important change was that the problem definition itself became finer-grained, allowing a degree of interpretability. Solutions to this finer-grained problem are already in the process of being integrated into a large-scale domain-specific search system that has had considerable impact in the last year on sex trafficking-related prosecutions.   

In keeping with the scope of this work
, we favor discussions of methodology over technical descriptions, except when necessary. Similarly, while we attempt to provide quantitative data when feasible, we also pay close attention to aspects of this work that cannot be easily controlled for or quantified, but still provide important lessons for deploying similarly complex systems.

\section{Related Work}
This work primarily draws on two emerging areas of research that are continuing to increase in significance. First, devising feasible and useful solutions for automatic trafficking detection is a good example of the general field of \emph{AI for social good}, at the levels of both algorithmic development and engineering effort. Second, the bias analysis conducted in this paper directly reflects recent debate on the (often unintentional) biases that creep into AI systems. Rather than attempt comprehensive coverage of these research areas, we focus on work that is closely tied to human trafficking (and where applicable, similar \emph{illicit} domains).

{\bf Intelligent Systems for Counter-Human Trafficking.}
Because of the alarming rise in online sex advertisement activity, a non-trivial portion of which may pertain to trafficking, building intelligent systems for assisting investigators and for enabling counter-human trafficking efforts has emerged as an important agenda. \cite{dubrawski2015leveraging} presented a host of data mining methods to support sex trafficking investigators in individual cases as well as lawmakers in understanding community-level statistics, amongst them the use of anomaly detection methods for community level statistics and an analysis of different text classifiers to detect sex trafficking-related activity. \cite{portnoff2017backpage} present techniques to link related escort advertisements including a stylometry based classifier as well as an approach that exploits data leakage in the payment system of advertisements conducted via bitcoin. \cite{nagpal2015entity} investigate clustering approaches for escort advertisements and rely on blocking schemes to link large amounts of escort advertisement data. Features such as rare n-grams or rare images can be used to create blocks of data within which exact comparisons of advertisements are carried out to generate clusters. The final cluster resolution is then achieved by subsequently resolving the dataset across blocks. 

More generally, there is a growing body of work for assisting investigative efforts in \emph{illicit domains}, which includes not just human trafficking (HT), but also domains such as illegal weapons sales and securities fraud, both of which have also been investigated under the Memex program. Often, both text and multi-modal data are involved. \cite{mattmann2016multimedia} explored metadata of multimedia files (such as Exif Tags in images and videos) retrieved from escort ad websites to assist HT forensics. \cite{gowda2017imageforensics} applied a deep learning-based computer vision framework to assist the detection and classification of ads related to illegal and dangerous weapons. Other relevant examples include building search systems for helping investigators search for, and research, promising leads. The Domain-specific Insight Graph (DIG) system is a good example of the latter and is currently being used by over 200 law enforcement agencies to counter human trafficking \cite{szekely2015:iswc,szekely2017}.


{\bf Trust and Bias in AI.}
A series of recent studies have shown that even \emph{standard} algorithms are not without bias. For example, \cite{caliskan2017semantics} show that standard learning algorithms trained on widespread text corpora learn stereotyped biases. 
\cite{sandvig2014auditing} provide an overview of research that criticizes and reverse engineers algorithms to understand consequences of their deployment and to discuss potential discrimination stemming from their use. 

The presence of undesired bias in feature representations automatically learned from the training dataset has also been studied in the literature; for instance, \cite{BCWS16} analyzed the gender biases in machine learning models.


The studies cited above make it clear that detecting bias in intelligent systems is both important and non-trivial. Thus, in describing our trafficking detection approach, we also present a \emph{bias mitigation plan} that emerged from months of effort, and key elements of which are already being integrated in an in-use counter-trafficking search system.    


\section{Problem Definition and Challenges}

We assume a data collection process that yields a domain-specific collection $C$ of webpages, almost all of which may be assumed to be either advertising sex (an \emph{escort ad}) or reviewing the services of an escort (a \emph{review ad}), collected via specially-tuned Web crawlers. In the most general case, each ad $c \in C$ is simply an HTML page, but it is convenient to assume that some preprocessing has been done (most importantly, \emph{text scraping} and \emph{information extraction}) and that $c$ is itself a set of \emph{key-value} pairs. More concretely, the \emph{schema} of $C$, which is a union over all keys in the collection, includes such attributes as phone numbers, locations, dates, and cleaned ad text, to name a few, and were used both for modeling and bias mitigation as well as in search systems that were eventually exposed to law enforcement.     

Given a small set $C' \subset C$ of ads from this collection, we define the \emph{automatic trafficking detection problem} as discovering an \emph{assignment} function $f: C' \rightarrow [0,1]$, where we denote $f(C')$ as the trafficking \emph{risk score} of the \emph{case study} $C'$. There are reasons for using this terminology. First, the manner in which $C'$ is isolated from $C$ is qualitatively similar to how a case study file is constructed. Ads in $C'$ are not sampled at random, although the precise reason for grouping the ads in $C'$ together may not be known in advance. Second, we note that any outputs by $f$ cannot be validated (even by a human reading the ad) except through a real-world investigation. In this sense, the problem is different from ordinary text or cluster classification problems, and more similar to Information Retrieval problems that seek to assess `relevance.' For this reason, $f$ is only said to assign a \emph{risk} score.       

Given the success and pervasive use of commercial search platforms like Google, why is an automated solution to trafficking detection even necessary? It is not unreasonable to assume a hypothetical \emph{lead generation} workflow whereby an investigator, after a period of browsing or exploration on high-activity portals like the \emph{adult} section of \url{backpage.com}, could make informed investigative decisions in an online fashion on whether a generated lead warrants a high-priority investigation. 

There are several problems that arise with such exploratory lead generation. First, although a sizable portion of trafficking activity takes place on the public Web, it is \emph{sparsely} interspersed among escort ads and reviews. In general, the connection between human trafficking and online sex advertising is still not well-understood \cite{latonero2011} but preliminary studies conducted under Memex provided some evidence that trafficking-related lead generation is a non-trivial problem. 
For \emph{investigative} purposes, non-trafficking leads do not take priority over those that may exhibit some form of trafficking. The second problem, which aggravates the effects of sparsity, is prohibitive \emph{scale}\footnote{The current Memex repository indexes over 100M documents.}. Although potential case study clusters can be isolated from this collection through a conservative process of rule- or keyword-based application, the many thousands of clusters that emerge as a result are far too many for a human to classify or study. Even more importantly, the domain is notoriously \emph{dynamic} and \emph{adversarial}, since potentially indicative signals (such as tattoos on specific body parts, and language use in advertisements) are constantly evolving, and are not well-understood sociologically.

Beyond addressing general problems of scale and sparsity (particularly, class skew), timely and automatic detection of trafficking-related activity has two other domain-specific motivations, both of which have the potential for widespread social impact. First, identifying cases at high risk can lead to early investigations, which may stop others from being trafficked. In this case, the motivation has a preventive aspect to it. Second, the ads, if flagged in time and retained \emph{offline}, have proven to be potent exhibits in actual trafficking-related criminal cases that were recently prosecuted in the US in New York and California \cite{greenemeier_2016}. By supporting prosecutions and evidence gathering, automatic detection systems have real potential for accelerating social justice and for raising the barrier to entry for traffickers that are using the Web to advertise illicit activity. 

\section{Approach}\label{sec:approach}

\begin{figure*}[t]\label{flowchart}
\includegraphics[height=3.54in, width=6.8in]{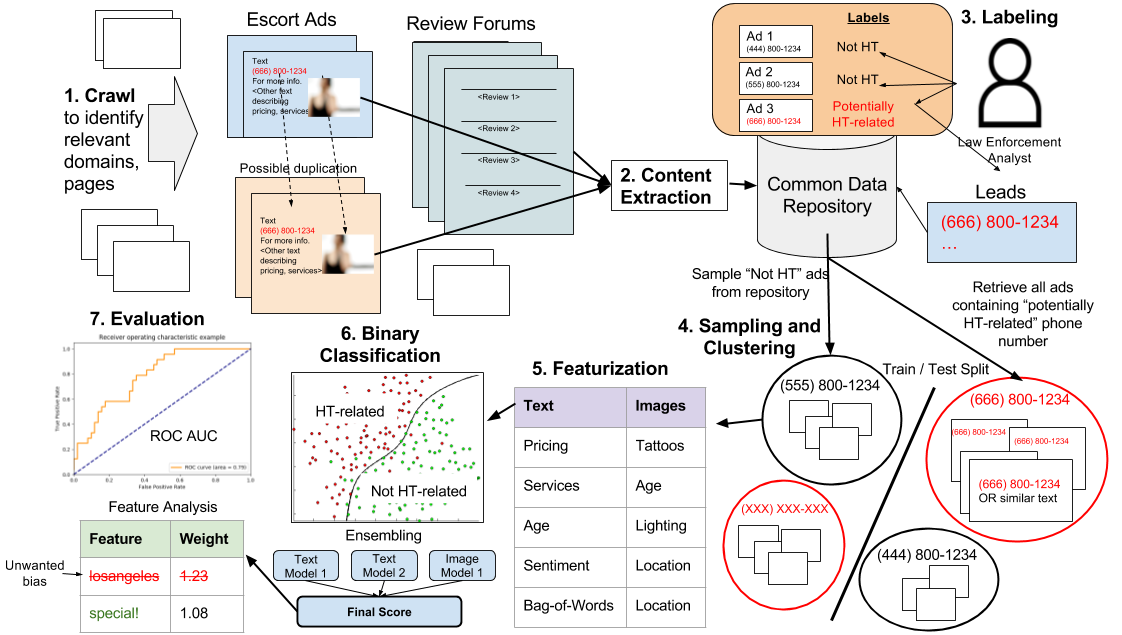}
  \caption{
  The end-to-end trafficking detection pipeline designed for multi-modal Web-scale corpora containing sex advertisements and reviews. This pipeline includes data collection stages such as \textit{crawling} and \textit{extraction}, data preparation stages such as \textit{sampling} and \textit{clustering} for acquiring labeled data for training and evaluation, and finally, the standard machine learning stages of \textit{featurization, training, ensembling, evaluation} and \textit{post hoc feature analysis}.}
\end{figure*}

Human trafficking detection, from the algorithmic perspective, is a binary classification task on retrieved escort ad and review data. Several tasks involved in this pipeline were performed as a collaborative effort among multiple research groups funded by the Memex program. Figure 1 illustrates the stages involved in an end-to-end pipeline designed for training and evaluating automatic trafficking detection:

\textbf{Step 1: Crawling and Data Collection:}
Web crawlers designed under Memex were tuned to search for, and scrape, sex-related advertisements, particularly from online marketplaces with significant \emph{escort-} and \emph{escort review-}related activity. Data analysis showed the majority of ads and reviews to be respectively coming from the \emph{adult} sections of \textit{backpage.com} and \textit{craigslist.com}, and \textit{eroticreview.com}. 
Data collected from crawlers was made available by indexing to the \textit{common data repository (CDR)} system. 

\textbf{Step 2: Extraction:}
In this stage, the crawled webpages and multimedia files were processed by \emph{Information Extraction} algorithms \cite{ie} that extracted domain-specific attributes such as 
phone numbers, dates, locations of services, image links, and plain text decriptions \cite{szekely2015:iswc,mattmann2011tika}. All extractions were indexed in the CDR along with the raw data.

\textbf{Step 3: Labeling:}
Labeling of escort data by trafficking experts is a costly and time-consuming task, involving significance cognitive burden. Several law enforcement partners across the U.S. shared information on historical and recent cases of (suspected) human trafficking that allowed us to retrieve relevant positive advertisements using unique identifiers such as phone numbers. Obtaining a reasonable amount of negatively labeled (i.e. not of investigative interest) ad data proved to be more difficult. Athough the Memex program contracted experts to acquire a training set of negatively labeled ads, the data was limited. 

\textbf{Step 4: Sampling and Clustering:}
\emph{Sampling} attempts to address the scarcity of negatively-labeled ads by relying on the assumption that the \emph{relative} prevalence of positive ads is small. The key idea is to obtain noisy negative labels by selecting groups of related advertisements \emph{at random} from the entire corpus, and labeling them as negative. However, since this process of obtaining negative labels is substantially different from the process of obtaining positive labels, the approach resulted in biased training subsets. This led to the development of a more rigorous negative sampling approach detailed further in the \textit{Bias Mitigation Plan}.

In the real world, as discussed in earlier section, ads are not independently created and there exist \emph{underlying clusters of ads} (denoted earlier as \emph{case studies}) that are generated by \emph{distinct entities} (such as escorts) but tend to contain \emph{similar} text and media by way of a \emph{latent relationship}. Clusters acquired for the purposes of training and evaluating the system need to satisfy downstream independent and identically distributed (i.i.d.) data modeling assumptions. To achieve this, we designed a set of multi-modal similarity functions over extracted attributes such as descriptive text, phone numbers, and images. Intuitively, these similarities can be used to cluster similar ads together and dissimilar items apart. 

\emph{Correlation clustering} is one such clustering approach that was found to be very useful in practice. There exist fast correlation clustering approaches with provable guarantees that scale linearly with the data, such as KWIKCLUSTER \cite{ailon2008aggregating}, which obtains a 3-approximation ratio. Parallel correlation clustering approaches \cite{pan2015parallel} based on KWIKCLUSTER have proven to be both efficient and effective in practice. Building on prior research findings \cite{elsner2009bounding}, we tuned and implemented a combination of KWIKCLUSTER, consensus clustering and local heuristics in our approach. 

\textbf{Steps 5 and 6: Featurization and Binary Classification}
Supervised text classification generally involves the mapping words and phrases into numerical feature vectors, using both classic bag-of-words approaches, and in more recent work, \emph{word embedding} algorithms like word2vec and fastText ~\cite{bojanowski2016enriching}. A variety of approaches for both `featurization' and model selection were explored throughout the project, but bias issues made systematic comparisons and validation of models a difficult task. We note that `black box' machine learning models such as deep neural networks provide no assurance that the learned features are not fitting to undesired biases in the labeled dataset (in addition to offering limited performance benefits for smaller training sets like ours \cite{NIPS2015_5782}). This led us to favor more transparent approaches, allowing for greater visibility into potential sources of bias in our system. Models that generally performed well, and offered (at least marginal) interpretability were linear-kernel SVMs (which are capable of suggesting the relative importance of features per \cite{guyon2002gene}), ensemble models such as random forest classifiers, and penalized logistic regression models trained on bag-of-word vectors. These modeling experiments informed both the \textit{Bias Mitigation Plan}, and the subsequent integration of the approach into end-user tools.
\textbf{Step 7: Evaluation:} The approach was evaluated on an independent set of ad clusters that was mutually exclusive from the training set, but was gathered using a similar protocol. We used the area under the Receiver Operating Characteristic (ROC) curve as the performance metric. Posthoc evaluation studies involved detailed analysis of feature importance, and the origin of important features. 

\section{Bias Mitigation Plan}

\begin{table*}
\centering
\caption{Bias Mitigation Overview}
\label{bias-plan}
\begin{tabular}{@{}llll@{}}
\toprule
\textbf{Bias Type} & \textbf{Overview} & \textbf{Diagnosis and Evaluation} & \textbf{Mitigation Steps} \\ \midrule
\multicolumn{1}{|l|}{Labeling}        & \multicolumn{1}{l|}{\begin{tabular}[c]{@{}l@{}}Labeled data biased toward\\ certain locations, Web \\ domains, and  positive class \\ due to nature of labeling \end{tabular}}                                     & \multicolumn{1}{l|}{\begin{tabular}[c]{@{}l@{}}1. Use correlation and hypothesis \\ tests to evaluate independence \\ of potentially-biased  features \\ relative to class\end{tabular}} & \multicolumn{1}{l|}{\begin{tabular}[c]{@{}l@{}}1. Sample additional negative \\ examples conditioned on biases \\ found in positive class data\\ 2. Remove biased  features\end{tabular}}                                                                                                                                     \\ \midrule
\multicolumn{1}{|l|}{Domain-Specific} & \multicolumn{1}{l|}{\begin{tabular}[c]{@{}l@{}}Cluster sizes vary and\\ escort ad content is \\ often duplicated across \\accounts and domains\end{tabular}}                                                                           & \multicolumn{1}{l|}{\begin{tabular}[c]{@{}l@{}}1. Single-feature modeling \\ (e.g. cluster size)\\ 2. Test for duplicate data \\ across classes\end{tabular}}                                                                                                                                                          & \multicolumn{1}{l|}{\begin{tabular}[c]{@{}l@{}}1. Sample negative clusters to \\ resemble positive cluster sizes \\ 2. Use multi-objective clustering to \\ prevent duplicated content from \\ appearing across clusters\end{tabular}}                                                                                           \\ \midrule
\multicolumn{1}{|l|}{Estimation}      & \multicolumn{1}{l|}{\begin{tabular}[c]{@{}l@{}}Training data is limited and \\ careless partitioning can \\ cause overfitting to samples \\ and invalid results\end{tabular}} & \multicolumn{1}{l|}{\begin{tabular}[c]{@{}l@{}}1. Classes in cross-validation \\ folds should show homogeneity \\ between  distributions of biased \\ features\end{tabular}}                & \multicolumn{1}{l|}{\begin{tabular}[c]{@{}l@{}}1. Condition cross-validation folds \\ to have matching distributions of \\ unwanted features\\ 2. Maintain model interpretability \end{tabular}} \\ \bottomrule
\end{tabular}
\end{table*}
\vspace{-0.1in.}


Mitigating bias in intelligent systems is a complicated issue, as the sources of bias are not easy to isolate. Bias may arise when certain algorithmic assumptions are violated, or when the training data is biased, either because the \emph{sample size} (compared to the population) is too small, or because the \emph{labeled data acquisition process} is biased. 

For example, while positive labels for groups of escort ads may come from a small number of law enforcement contacts that only provide cases for specific regions,  noisy negative labels sampled at random will follow the true location distribution in the corpus more closely. Classifiers trained on such data may learn to differentiate classes using locations specific to law enforcement contacts, rather than learn actual and meaningful signals indicating human trafficking.





\subsection{Diagnosing and Evaluating Bias}
The first step in addressing bias is identifying potential sources of bias that often become apparent through thorough data exploration, domain understanding, or the modeling process. Simple statistical significance tests or the computation of correlation coefficients can assist in validating suspected biases. Also, expert knowledge can also help us determine obvious information in the data that should not be indicative of the output label.

For example, we can test the null hypothesis that the distribution of a potentially-biased feature like \textit{Web domain} is independent with respect to label class (i.e. not biased via sampling, labeling, clustering, etc.) with a Pearson's chi-squared test \cite{pearson1900x}.
Using $ \alpha = 0.05 $ as the threshold for rejecting our null hypothesis, the below table shows actual counts of ads for labeled data available for modeling during the program:
\vspace{-0.2in.}
\begin{table}[H]
\centering
\caption{Ads by Domain Group and Label Class}
\label{my-label}
\begin{tabular}{@{}|l|l|l|l|@{}}
\toprule
               & Positive         & Negative         & \textbf{Total}   \\ \midrule
backpage.com   & 165,686          & 125,467          & \textbf{291,153} \\ \midrule
other          & 155,271          & 154,627          & \textbf{309,898} \\ \midrule
\textbf{Total} & \textbf{320,957} & \textbf{280,094} & \textbf{601,051} \\ \bottomrule
\end{tabular}
\end{table}
Calculation of the test statistic and subsequent p-value 
results in $p < 0.00001$ and our assumption of independence is violated at the chosen $\alpha$. To correct for this bias we can sample additional (presumably) negative ads from the CDR with the aim of aligning the two distributions. Although these are actual numbers, this is a simplified example intended for demonstration -- a more appropriate correction would involve more granular groupings of domains and other biased features would need to be considered during the sampling of additional negatives.

Under a more traditional multiple hypothesis testing scenario this would necessitate correcting for the issue of multiple comparisons using, for instance, a Bonferroni correction \cite{curran2000multiple} to set a more rigorous threshold for finding statistically significant relationships. Our motives differ, however, in that we are not using hypothesis tests to identify variables that may help explain a resulting dependent variable, but rather that we are using domain knowledge to determine information that should not strongly correlate with the class labels.


In addition to testing for bias, this approach can be used to evaluate the effectiveness of mitigation efforts described below. It should also be used when constructing folds for cross validation to ensure independence assumptions hold during learning. If mitigation is successful independence assumptions should be satisfied according to the above criteria.

\subsection{Mitigation of Bias}
\textbf{Information Removal and Conditioned Sampling.}
Upon identifying biased information in training data, we want to ensure that we correct for the bias before or during model training. One way to achieve this is to try to remove any information that relates to a biased feature either before or after vectorization of the data points. For example, in the context of \textit{location bias}, one could try to remove any tokens in the ad text that refer to locations. Naturally, this requires information extractors with high recall and ideally high precision, which may not always be available. 

Another approach towards alleviating this issue is to add or remove data points from the training data with the goal of ensuring that the distribution of the biased feature is similar across the positive and negative data points. Adding additional human-labeled data, especially in such a way as to align distributions is difficult, expensive, and may introduce additional unwanted sampling biases. 
In this context, a more appropriate method to control for bias is to extract additional negative ads from the CDR such that our sample contains similar distributions of biased features across classes, discouraging classifiers from learning them. As mentioned in \textit{Step 4} of the \textit{Approach} section, the process of sampling of negative ads and clusters from the CDR is conducive to the conditioning, especially when extractions for biased features are readily available. This strategy introduces the step of determining how to measure similarity between distributions, which may be done via distribution divergence measures such as the R\'enyi divergence \cite{van2014renyi} or by doing a two-sample Kolmogorov--Smirnov test~\cite{daniel1978applied}.  




\textbf{Clustering.}
Ignoring \textit{inherent} dependencies between ads introduces numerous problems when the aim is to build text classifiers that assume i.i.d. data. For example, the implicit weighting of features that occurs in a dataset with many near duplicates will not reflect the importance of the features in relation to the output label one is trying to predict.  Rather, a model will be encouraged to memorize specific patterns in training data duplicated across classes. This is especially problematic under widely varying cluster sizes, which would suggest \emph{cluster-level classification} as a sensible scheme.

Recovering the true underlying clusters presents unique challenges. The same person or group may produce ads for multiple individuals. Additionally, ad text and images are sometimes copied across personas and phone numbers are often intentionally obfuscated. As described in \emph{Step 4} of the \emph{Approach}, we used a correlation clustering method and some local heuristics to achieve subjectively good clustering results. To estimate how well the clustering helped in creating a clean training and test split as well as clean cross-validation folds one can estimate the out-of-cluster loss of a model under dependency within latent clusters, i.e. a way to recover the loss in the independent setting, such as shown in \cite{Barnes2017}. %

\textbf{Indicator Mining and Integration.}
In Table 1, interpretability of the model is a key mitigation step and involves a social aspect since the utility of interpretability is to the \emph{users} of the system. Detailed conversations with law enforcement officials revealed a strong desire that the systems produce finer-grained `clues' suggestive of potential, context-dependent trafficking detection, rather than a single score. These clues, called \emph{indicators}, are highly specific and are designed to detect such high-level features as \emph{escort movement}, advertisement of \emph{risky} sex services, and presence of \emph{multiple girls} within a single advertisement and several others. Indicators are defined to be features that are (believed to be) relevant to inferring accurate trafficking risk scores. In the last phase of the program, we significantly extended the principles of the approach with expert-elicited rules and unsupervised text embeddings to \emph{supplement} ads with indicators.    
At the time of writing, these indicators are actively being integrated into the DIG search system, currently in use by more than 200 U.S. law enforcement agencies. 


\section{Impact and Conclusion}
The majority of Memex trafficking detection systems are being permanently transitioned to the office of the District Attorney of New York, and generic `non-trafficking' versions have been released as open-source software in the DARPA Memex catalog\footnote{\url{https://opencatalog.darpa.mil/MEMEX.html}}. In the last year, DIG, along with other trafficking detection tools from Memex, has led to at least three trafficking prosecutions, including a recently concluded case in San Francisco where a man was sentenced to 97 years to life for human trafficking\footnote{\url{http://www.sfgate.com/crime/article/
   Man-sentenced-to-97-years-in-human-\\   trafficking-7294727.php}}. More than 25 victims were rescued, and the DA's office in San Francisco publicly acknowledged the Memex tools in making this possible. 

This paper presented and defined an important problem called trafficking detection, which has much potential to be aided by recent advances in intelligent systems. We presented a general approach to the problem developed and evaluated over years of research under the DARPA Memex program, and a mitigation plan for addressing biases in the approach. Given Memex's sustained impact, we hope to continue improving our trafficking detection systems.  

{\bf Acknowledgements.}This effort was supported in part by JPL, managed by the California Institute of Technology on behalf of NASA, and additionally in part by the DARPA Memex/XDATA/D3M programs and NSF award numbers ICER-1639753, PLR-1348450 and PLR-144562 funded a portion of the work. 
The authors particularly thank Dr. Chris Mattmann from JPL, Dr. Pedro Szekely from ISI, and Dr. Artur Dubrawski from Carnegie Mellon University for their support and all of the Memex collaborators for their contributions.

\balance
\bibliographystyle{aaai}
\bibliography{references}

\end{document}